\documentclass[lettersize,journal]{IEEEtran}
\usepackage{amsmath,amsfonts}
\usepackage{algorithmic}
\usepackage{algorithm}
\usepackage{array}
\usepackage[caption=false,font=normalsize,labelfont=sf,textfont=sf]{subfig}
\usepackage{textcomp}
\usepackage{stfloats}
\usepackage{url}

\usepackage{booktabs}
\usepackage{graphicx}
\usepackage{amssymb}
\usepackage{cite}
\usepackage{multicol}
\usepackage{multirow}
\usepackage[table]{xcolor}
\usepackage{amsthm} %proof

\usepackage{verbatim}

\begin{document}

\title{PAL: Prompting Analytic Learning with Missing Modality for Multi-Modal Class-Incremental Learning}

\author{
Xianghu Yue,
Yiming Chen,
Xueyi Zhang,
Xiaoxue Gao,
Mengling Feng,
Mingrui Lao,
Huiping Zhuang,
Haizhou Li,~\IEEEmembership{Fellow,~IEEE}
        % <-this % stops a space
\thanks{
Xianghu Yue and Yiming Chen are with the Department of Electrical and Computer Engineering, National University of Singapore, Singapore 117583.
(email: xianghu.yue@u.nus.edu, yiming.chen@u.nus.edu).}
\thanks{
Xueyi Zhang and Mingrui Lao are with the Laboratory for Big Data and Decision, National
University of Defense Technology, Changsha, 410073, China. (email: zhangxy1998@nudt.edu.cn, laomingrui17@nudt.edu.cn)}
\thanks{Xiaoxue Gao is with Institute for Infocomm Research, A*STAR, Singapore 138632.
(email: Gao\_Xiaoxue@i2r.a-star.edu.sg).}
\thanks{Mengling Feng is with Saw Swee Hock School of Public Health, National University of Singapore, Singapore 117583. (email: ephfm@nus.edu.sg)}
\thanks{Huiping Zhuang is with Shien-Ming Wu School of Intelligent Engineering, South China University of Technology Guangzhou, Guangdong, 510006, China. (email: bruis\_zhuang@hotmail.com)}
\thanks{Haizhou Li is with Shenzhen Research Institute of Big Data, and School of Data Science, The Chinese University of Hong Kong, Shenzhen, 518172, China;
also with the Department of Electrical and Computer Engineering, National
University of Singapore, 117583.
(email: haizhouli@cuhk.edu.cn).}
}

\markboth{Journal of \LaTeX\ Class Files,~Vol.~14, No.~8, August~2021}%
{Shell \MakeLowercase{\textit{et al.}}: A Sample Article Using IEEEtran.cls for IEEE Journals}


\maketitle

\begin{abstract}
Multi-modal class-incremental learning (MMCIL) seeks to leverage multi-modal data, such as audio-visual and image-text pairs, thereby enabling models to learn continuously across a sequence of tasks while mitigating forgetting.
While existing studies primarily focus on the integration and utilization of multi-modal information for MMCIL, a critical challenge remains: the issue of missing modalities during incremental learning phases.
This oversight can exacerbate severe forgetting and significantly impair model performance.
To bridge this gap, we propose PAL, a novel exemplar-free framework tailored to MMCIL under missing-modality scenarios.
Concretely, we devise modality-specific prompts to compensate for missing information, facilitating the model to maintain a holistic representation of the data.
On this foundation, we reformulate the MMCIL problem into a Recursive Least-Squares task, delivering an analytical linear solution.
Building upon these, PAL not only alleviates the inherent under-fitting limitation in analytic learning but also preserves the holistic representation of missing-modality data, achieving superior performance with less forgetting across various multi-modal incremental scenarios.
Extensive experiments demonstrate that PAL significantly outperforms competitive methods across various datasets, including UPMC-Food101 and N24News, showcasing its robustness towards modality absence and its anti-forgetting ability to maintain high incremental accuracy.
\end{abstract}

\begin{IEEEkeywords}
multi-modal, missing modality, class-incremental learning.
\end{IEEEkeywords}

\section{Introduction}
\IEEEPARstart{O}{ur} daily observations are inherently multi-modal, encompassing acoustic, visual, and linguistic signals, thus modeling and coordinating multi-modal information is of great interest and offers broad application potential.
By leveraging the complementary properties of different modalities, such as audio-visual or image-text pairs, these endeavors have shown to enhance the understanding and capabilities of models for tasks such as speech recognition~\cite{afouras2018deep, burchi2023audio, tao2020end, yeo2024akvsr}, visual recognition~\cite{wu2023revisiting, kunpeng2023}, and sound localization~\cite{qian2022deep,qian2022audio,chen2021localizing,chen2022sound,hu2019deep,senocak2018learning}.
% To be mentioned, in real-world applications, it is crucial for models to adapt to evolving sets of classes and integrate new information into an ever-expanding knowledge base.

\begin{figure}
    \centering
    \includegraphics[scale=0.22]{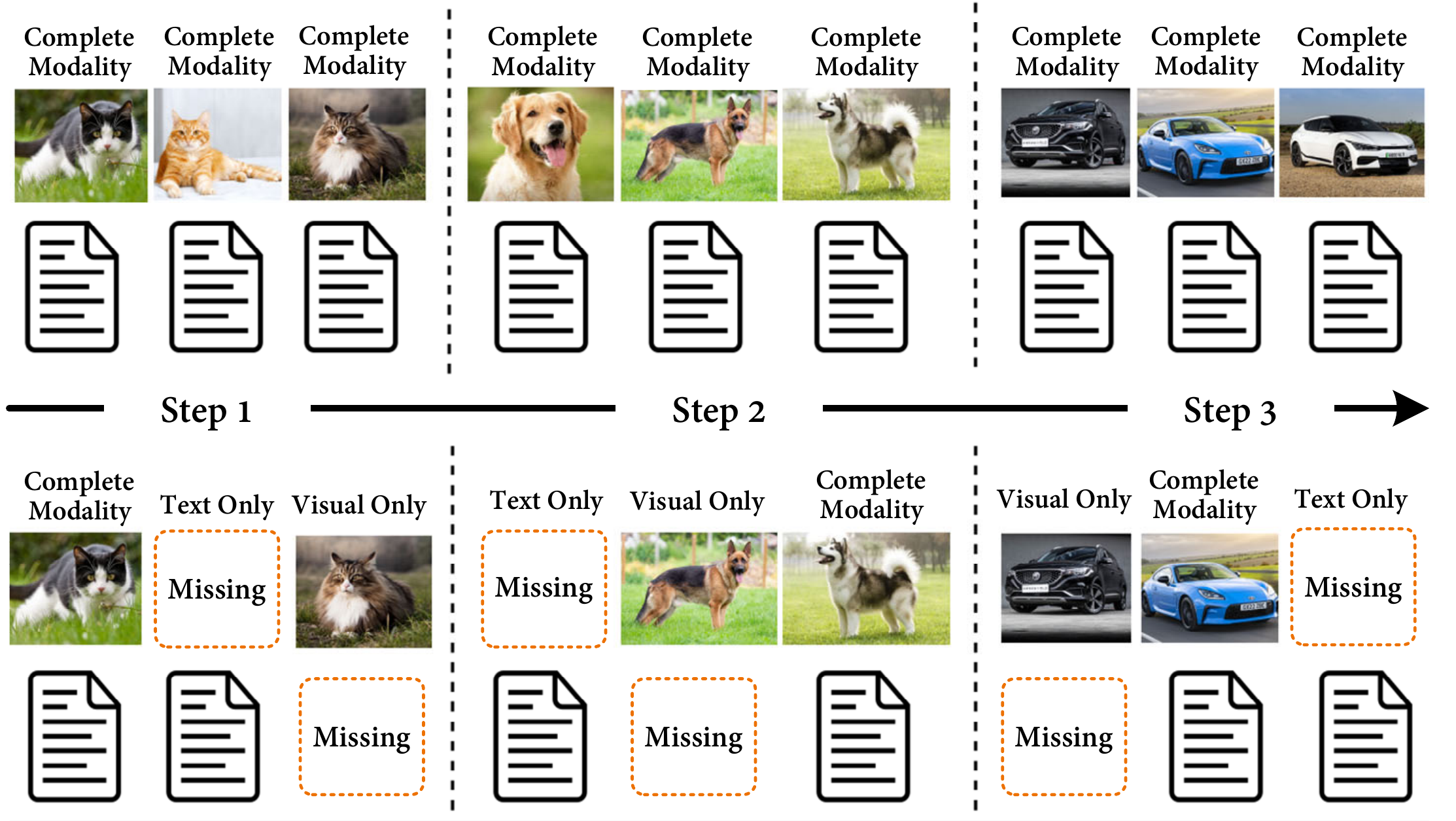}
    \caption{Illustrate of multi-modal class-incremental learning (MMCIL) with missing modality. Previous works (top) focus on leveraging multi-modal information from modality-complete data to mitigate forgetting. In contrast, our work (bottom) studies a more general scenario where various modality-missing cases would occur differently not only across each data sample but also at various incremental learning phases.}
    \label{fig:intro}
    % \vspace{-0.2cm}
\end{figure}

In real-world applications, it is crucial for models to adapt to evolving set of classes and integrate new information into an expanding knowledge base.
Class-incremental learning (CIL)~\cite{rebuffi2017icarl, li2017learning} addresses this need by progressively updating a network's parameters as new classes are introduced over time,
% Class-incremental learning (CIL)~\cite{rebuffi2017icarl, li2017learning} progressively updates the network's parameters by incorporating new training data from different unseen classes over time, 
which is intuitively inspired by the human learning process, where individuals assimilate new information continuously on top of previously-acquired knowledge.
Existing techniques mainly employ rehearsal-free~\cite{kirkpatrick2017overcoming, li2017learning, jung2016less} and replay-based~\cite{zhang2020class,douillard2020podnet,liu2021adaptive,liu2021rmm,liu2020mnemonics} solutions to mitigate the challenge of catastrophic forgetting.
% Recently, a novel technical route to CIL, the Analytic Learning (AL)-based~\cite{zhuang2022acil, Zhuang2024} techniques, addressing the issue of forgetting by identifying the iterative mechanism as the main cause and substituting it with linear recursive tools, have shown promising results.
Recently, a novel technical route to CIL, the Analytic Learning (AL)-based~\cite{zhuang2022acil, Zhuang2024} techniques, has demonstrated promising results.
These techniques tackle the issue of forgetting by identifying the iterative mechanism as the main cause and substituting it with linear recursive tools.
Nonetheless, existing AL-based methods often face under-fitting limitations due to their reliance on a frozen backbone, which lacks the ability to update dynamically with new classes over time.
% their reliance on a single linear projection and the frozen backbone.

While prior CIL works have primarily focused on single modality, recent efforts~\cite{mo2023class, Pian_2023_ICCV, zhao2024arxiv, mmal} have extended CIL into broader domains and modalities, giving rise to multi-modal class-incremental learning (MMCIL).
MMCIL seeks to leverage multi-modal data to enable models to learn a sequence of tasks incrementally while addressing the problem of catastrophic forgetting.

However, existing MMCIL methods typically require the completeness of all modalities during both training and evaluation, limiting their applicability in real-world scenarios where subsets of modalities may be missing during incremental phases (see Figure~\ref{fig:intro}).
The uncertain absence of modalities would lead to more severe forgetting and significantly degrade the model performance, particularly during CIL procedures~\cite{zhao2024arxiv, MMIN2024guo}.

In this paper, we explore the MMCIL under a more practical yet challenging modality-incomplete scenario, where various miss-modality cases may occur for each sample during the CIL training and test phases.
To this end, we propose a novel Prompting Analytic Learning (PAL) framework, PAL, stemming from prompt learning and analytic learning with a dual-module architecture.
On the one hand, the first module leverages modality-specific prompts to compensate for missing information, facilitating the model to maintain a holistic representation of the data.
On the other hand, the second module reformulates the MMCIL problem into a Recursive Least Squares (RLS) task, delivering an analytical solution.
Building upon these, PAL not only alleviates the inherent under-fitting limitation in analytic learning but also preserves the holistic representation of missing-modality data, achieving better performance with less forgetting across various multi-modal incremental scenarios.

The contributions of our paper can be summarised as follows:
\begin{itemize}
    \item We propose PAL -- a novel exemplar-free technique to address the missing modality problem during the multi-modal CIL procedure.
    \item PAL seamlessly incorporates modality-specific prompts into analytic learning, redefining the MMCIL with missing modality into an RLS task and resolving the intrinsic under-fitting limitations via prompt tuning.
    % \item By leveraging the prompt-based tuning, PAL addresses the intrinsic under-fitting issue associated with analytic learning, enhancing the model's overall capability.
    \item Experimental results demonstrate that PAL outperforms recent state-of-the-art methods, showcasing its robustness in addressing the missing modality problem and its ability to maintain high classification accuracy during CIL procedures.
\end{itemize}

\section{Related work}
\subsection{Multi-Modal Learning with Missing Modalities}
Multi-modal learning methods exploit the complementary information from different modalities, (e.g., image, text, or audio) to jointly represent a common concept~\cite{Tadas2019tpami, Dhanesh2017}.
Recent multi-modal transformers~\cite{vilt2021, vatt} emerge as unified models capable of processing inputs from different modalities and fuse them through token concatenation without relying on modality-specific feature extractors.
These transformer-based models have been widely applied in various multi-modal tasks, such as visual question answering, video captioning, and cross-model retrieval~\cite{gabeur2020mmt, vilt2021, Botach2022}.
However, most multi-modal learning methods assume the availability of all modalities, which is often not feasible in real-world scenarios, posing a significant challenge for multi-modal learning.
When one of the modalities is missing, the multimodal fusion becomes unreachable and the model might predict inaccurately~\cite{mengmeng2022cvpr, Lee2023cvpr}.
Several studies~\cite{mengmeng2022cvpr, woo2023, zhao2021} have explored to build multi-modal models that are robust to missing-modality data.
MMIN~\cite{zhao2021} predicts the representation of any absent modality based on available ones through a joint multi-modal representation, while SMIL~\cite{woo2023} estimates latent features of the modality-incomplete data via Bayesian meta-learning.

Recently, the use of prompt tokens within pre-trained transformers to represent the information of missing modalities has gained considerable interest~\cite{Lee2023cvpr, Jang2024icassp, wang2022cvpr, wang2022eccv, zhao2024arxiv}, due to its ability to effectively encapsulate modality information with limited resources. 
However, these approaches are typically trained and tested on fixed datasets, which limits their effectiveness in handling dynamic or evolving data common in real-world applications.
In contrast, in this work, we target to address the missing-modality challenge in dynamic and class-incremental learning scenarios, where data and classes evolve over time.
% We utilize prompt tokens to invoke the intrinsic knowledge within pre-trained models, subsequently rendering them into concise surrogates for the missing modalities, which not only streamlines the representation of absent information but also capitalizes on knowledge in models, thereby enhancing the robustness and adaptability of models in scenarios confronted with missing modalities.

\subsection{Class-Incremental Learning}
The core objective of CIL is to enable models to acquire new knowledge without forgetting what they have previously learned.
This field can be broadly categorized into four categories: regularization-based, memory-based, prompt-based, and analytic learning-based methods.

\subsubsection{Regularization-based} These methods introduce constraints to the learning process to minimize the impact of new tasks on previously learned knowledge. 
For instance, EWC~\cite{kirkpatrick2017overcoming} protects the critical parameters associated with previous tasks, ensuring their stability and minimizing their susceptibility to change during training for new tasks.  
Similarly, LfL~\cite{jung2016less} penalizes differences in network activations, while LwF~\cite{li2017learning} prevents activation changes between old and new networks.

\subsubsection{Memory-based} Memory-based techniques ~\cite{bang2021rainbow,rebuffi2017icarl,zhao2021memory, castro2018end} seek to tackle the forgetting issue by utilizing and storing data from previous tasks.
This mechanism was first introduced by iCaRL~\cite{rebuffi2017icarl}, which inspired various subsequent approaches due to its superior effectiveness.
PODNet~\cite{douillard2020podnet} employs a spatial-based distillation loss to preserve previously acquired knowledge while learning new information.
LUCIR~\cite{hou2019learning} introduces a novel adaptation technique that replaces the softmax layer with a cosine layer to better balance old and new classes.
Although these methods achieve strong performance, they still inherently need to store previous samples, imposing limitations on their feasibility.

\subsubsection{Prompt-based} Prompt-based methods represent a significant evolution in CIL approaches.
These methods use learnable prompts for each task to extract and apply knowledge from the frozen pre-trained backbones~\cite{wang2022cvpr, jung2023, wang2022eccv} without maintaining the memory buffer.
L2P~\cite{wang2022cvpr} firstly introduces the prompt pool to tune the frozen ViT backbone for CIL tasks.
Then, DualPrompt~\cite{wang2022eccv} utilizes two complementary prompts, G-Prompt and E-Prompt, to learn task-invariant and task-specific knowledge, respectively.

\subsubsection{Analytic learning-based methods} leverages the principles of least squares to achieve closed-form solution for network training~\cite{zhuang2021blockwise,guo2004pseudoinverse}.
ACIL\cite{zhuang2022acil}, for example, first transforms CIL into a recursive analytic learning process which releases the need of storing exemplars by preserving a correlation matrix. 
DS-AL~\cite{Zhuang2024} builds on this by employing a dual-stream approach to further enhance the fitting ability of ACIL.
The following work GKEAL~\cite{zhuang2023gkeal} introduces a Gaussian kernel process to specialize the few-shot CIL setting.

AL-based and prompt-based methods, as emerging branches of CIL, both have shown impressive performance. 
However, most existing works primarily focus on exploring single-modality, neglecting the potential benefits of multi-modal learning.
More importantly, the AL-based methods require freezing the backbone during the incremental steps, which will cause serious under-fitting limitations due to the complexity of multi-modal data, i.e., the missing-modality scenario, while the prompt-based techniques aim to insert prompts without training the pre-trained backbone, 
As a result, we are motivated to explore prompting analytic learning with the missing modality for MMCIL.

\section{Methodology}
The proposed MMCIL method, PAL, owns two main modules: a prompt module, $\mathcal{P}$, and an AL module, $\mathcal{A}$, where $\mathcal{A}$ has a multi-modal backbone, $\mathcal{F}$, and a classifier $\mathcal{C}$.
PAL incrementally trains $\mathcal{P}$ and $\mathcal{C}$ with backpropagation (BP) and then re-trains $\mathcal{C}$ with analytic learning (AL).
Before going into depth, we provide the problem definition and framework overview.

\subsection{Problem Definition}
Without loss of generality, we consider a multi-modal dataset with visual and text modalities.
In MMCIL setting, for an incremental step $\mathcal{T}_k$, its training and testing sets are: $\mathcal{D}_k^{\text{train}} \sim {\{X^{\text{train}}_{k, v}, X^{\text{train}}_{k,t}, Y^{\text{train}}_{k}}\}$,
$\mathcal{D}_k^{\text{test}} \sim {\{X^{\text{test}}_{k, v}, X^{\text{test}}_{k,t}, Y^{\text{test}}_{k}}\}$,
where $X_{k,v}$ and $X_{k,t}$ are the sample's visual and text modalities respectively, and $Y_{k} \in \mathcal{C}_k$ is the label class, where $\mathcal{C}_k$ is the label space of task $\mathcal{T}_k$.
For any two tasks' label space, $\mathcal{C}_{k1}$ and $\mathcal{C}_{k2}$, they are mutually exclusive, i.e., $\mathcal{C}_{k1} \cap \mathcal{C}_{k2} = \varnothing$.
The objective of MMCIL with missing modality at step $\mathcal{T}_k$ is to train the networks given $\mathcal{D}_k^{\text{train}}$, and test them on $\mathcal{D}_{1:k}^{\text{test}}$, despite one modality is missing.
The data sample may contain different missing cases including complete, text-only, and image-only data.
Following the prior work~\cite{Lee2023cvpr}, we assign dummy inputs (e.g., empty pixel/string for image/text) to the missing-modality data.

\subsection{Framework Overview}
\begin{figure}
\centering
\includegraphics[scale=0.83]{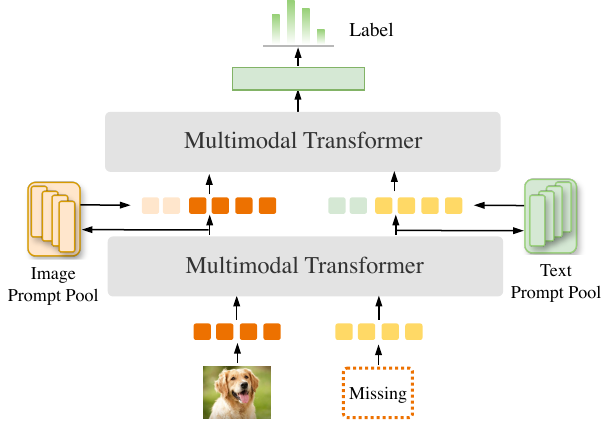}
% \vspace{-0.4cm}
\caption{The basic backbone of PAL, including a pre-trained multi-modal transformer, two modality-specific prompt pools (i.e., image and text), and a linear classifier.}
% \vspace{-0.2cm}
\label{fig:backbone}
\end{figure}

\begin{figure*}
\centering
\includegraphics[scale=1.05]{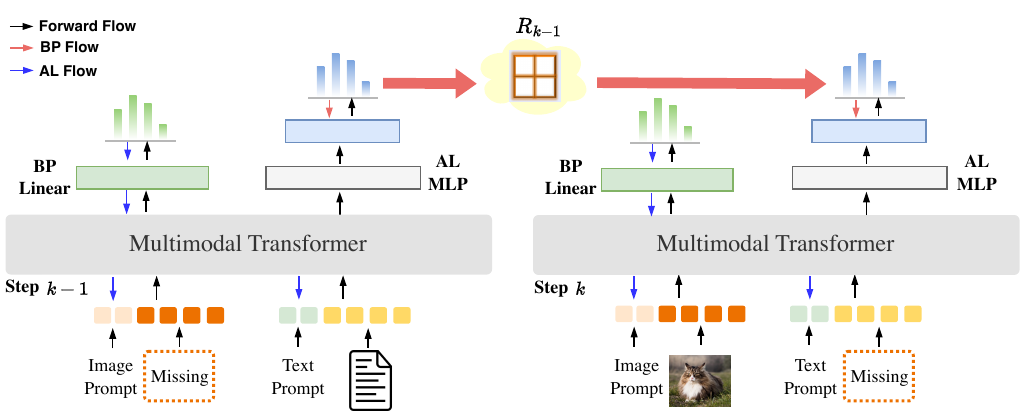}
    \caption{The conceptual illustration of our proposed PAL framework during multi-modal CIL procedures, consisting of two steps of training. The first step trains the classifier via BP, and the second step re-trains the classifier via AL.
    For simplicity, we omit the modality-specific prompt pool.
    }
    \label{fig:pal}
\end{figure*}

As illustrated in Figure~\ref{fig:backbone}, we use the pre-trained multi-modal transformer, i.e., ViLT~\cite{vilt2021}, for our multi-modal backbone $\mathcal{F}$.
The classifier $\mathcal{C}$ consists of a linear fully-connected layer during BP-based training, while one additional up-sampling linear layer will be inserted during AL-based re-training.
There are two modality-specific pools (i.e., $\mathcal{P}^v$ and $\mathcal{P}^t$) for the prompt module $\mathcal{P}$.
During each CIL procedure, we first train $\mathcal{P}$ and $\mathcal{C}$ via BP, then re-train $\mathcal{C}$ via AL for the final classification, as shown in Figure~\ref{fig:pal}.

\subsection{Modality-Specific Prompt}
For each modality, we create a modality-specific prompt pool to store the modality-specific knowledge, i.e., $P_v$ for image and $P_t$ for text.
Each pool contains learnable attention weights $A^{m}  \in \mathbb{R}^{D \times N}$, and keys $E^{m} \in \mathbb{R}^{D \times N}$, where $D$ is the dimension, $N$ is the size of the pool and $m$ represents a specific modality (either visual or text, i.e., $m \in \{v, t\}$. 
The prompt pool for each modality is denoted by $\mathcal{P}^{m} = \{ P^{m}_n \}_{n=1}^N$, where each prompt $P^{m}_n \in \mathbb{R}^{D \times N_{\mathrm{p}}} $ and $N_{\mathrm{p}}$ is the prompt length.

We mainly follow the prompt selection mechanism~\cite{Smith_2023_CVPR} to obtain prompts, in which
we first tokenize the image into a sequence of patch embedding 
$X^{v} = [v_1, \cdots, v_V]$, and the text input into a sequence of token embedding  $X^{t} = [t_1, \cdots, t_T]$.
Then we concatenate the $X^{v}$ and $X^{t}$ as well as  two special tokens $\mathbf{x}_{\text{cls}}^{t}$ and $\mathbf{x}_{\text{cls}}^{v}$ as the input of the multi-modal backbone $\mathcal{F}$, and generate an output sequence of the same length:
\begin{equation}
\label{eq:qe}
    [\tilde{\mathbf{x}}_{\text{cls}}^{t}, \tilde{X}^{t}, \tilde{\mathbf{x}}_{\text{cls}}^{v}, \tilde{X}^{v}] = \mathcal{F}([\text{x}_{\text{cls}}^{t}, X^{t}, \mathbf{x}_{\text{cls}}^{v}, X^{v} ]).
\end{equation}
We set $\tilde{\mathbf{x}}_{\text{cls}}^{v}$ as $\text{q}^{v}$, which is the query embedding for image and  $\tilde{\mathbf{x}}_{\text{cls}}^{t}$ as $\text{q}^{t}$, which is the query embedding for text.

The generated image/text query embeddings are further utilized to generate the weight vector $\mathbf{w}^{m} = [{w}^{m}_1, \cdots, {w}^{m}_N]$ ($m \in \{v, t\}$) as:

\begin{equation} \label{eq:weight-vector}
    \begin{aligned}
        w^{v}_n = sim(\text{q}^{v} \odot A^{v}_n, E^{v}_n),  \forall n \in [1,N],  \\
        w^{t}_n = sim(\text{q}^{t} \odot A^{t}_n, E^{t}_n), \forall n \in [1,N],  \\
    \end{aligned}
\end{equation}
where $sim(\cdot,\cdot)$ measures consine similarity, $\odot$ is Hadamard product, and $A^{v}_n/A^{t}_n$ denotes the $n$-th column of $A^{v}/A^{t}$. 

Finally, we use the computed weights $\mathbf{w}^{v}$ and $\mathbf{w}^{t} $ in Equation~\eqref{eq:weight-vector} to generate image and text prompts, respectively, via a weighted sum of the prompt components:
\begin{equation} \label{eq:weight-summation}
    \begin{aligned}
        P^{m} = \sum_{n=1}^N w^{m}_n \cdot P^{m}_n, \forall m \in \{v, t\}.
    \end{aligned}
\end{equation}

To fully utilize the modality-complete samples, we further construct its missing-modality counterparts by replacing the original input with the dummy text and image input to obtain the visual-only and text-only samples.
These missing-modality samples are then used to generate the reconstructed text query embedding $\hat{\text{q}}^{t}_i$ and the reconstructed visual query embedding $\hat{\text{q}}^{v}_i$.
Meanwhile, we obtain the ground-truth text and visual query embedding $\text{q}^{t}_i/\text{q}^{v}_i$ based on the original text and visual input as Equation~\eqref{eq:qe}. 
The reconstruction loss is computed by:
\begin{equation}
    \begin{aligned}
    \mathcal{L}_r = \frac{1}{L_c} \Big \{ \sum_{i=1}^{L_c} \|  \text{q}^{v}_i - \hat{\text{q}}^{v}_i \|_2^2 + \| \text{q}^{t}_i - \hat{\text{q}}^{t}_i   \|_2^2  \Big \}.
    \end{aligned}
\end{equation}
where $L_c$ is the number of modality-complete samples at $\mathcal{T}_k$.

% Note that, the modality-specific prompt learning relies on condition that a sample contains both textual content $\mathbf{t}$  and visual content $\mathbf{v}$ for generating text query embedding $\mathbf{q}^{\mathrm{t}}$ and visual query embedding  $\mathbf{q}^{\mathrm{v}}$.
% Nevertheless, as above mentioned, in real scenarios, we might have only access to the content of a single modality and fail to explicitly obtain both  $\mathbf{q}^{\mathrm{t}}$ and $\mathbf{q}^{\mathrm{v}}$. To address the modality-missing challenge, we proposed to reconstruct the query embedding of the missing modality based on a memory bank $\mathcal{M}$.
% We will introduce the detailed process of reconstructing the missing query embedding in Section~\ref{sec:mqr}.

\subsection{Prompting Analytic Incremental Learning}
PAL reformulates MMCIL into an RLS task, in which the multi-modal backbone $\mathcal{F}$ has to be frozen.
However, when the training samples are complex, particularly multi-modal data with missing modalities, the network may suffer from under-fitting to capture discriminative information between tasks~\cite{zhuang2022acil, ds_al}.
To alleviate this issue, we prepend the above-introduced modality-specific prompts to the input and dynamically tune the prompt $\mathcal{P}$ and classifier $\mathcal{C}$ on new samples via BP first.
After which, we re-train $\mathcal{C}$ via AL for the final prediction.

\subsubsection{Prompt tuning via BP}
We first train the prompt module $\mathcal{P}$ and the classifier $\mathcal{C}$ with an iterative BP for multiple epochs on training data $\mathcal{D}_{k-1}^{\text{train}}$ at step $k-1$.
Specifically, the objective is jointly optimized on two loss functions:
\begin{equation}
\label{eq:bp_train}
    \mathcal{L} = \mathcal{L}_c + \lambda \mathcal{L}_r
\end{equation}
where $\mathcal{L}_c$ is the cross-entropy loss for classification, and $\lambda$ is a reconstruction weight to balance the two loss values.

After the BP training, given a multi-modal input sample $(X_{k-1,v}^{\text{train}}, X_{k-1, t}^{\text{train}})$, the network produces the following output:
\begin{equation}
    Y = f_{\text{softmax}} (\mathcal{F}([P^t, X_{k-1, t}^{\text{train}}, P^v, X_{k-1,v}^{\text{train}}]) W_{\text{FNN}})
\end{equation}
where $W_{\text{FFN}}$ are the parameters of the linear feed-forward classifier, and $f_{\text{softmax}}$ denotes the softmax function.

Upon completing the prompt-based tuning, PAL seeks to detach the frozen multi-modal backbone with a tuned prompt pool and attach it with a 2-layer AL network for re-training, in which the first layer conducts representation up-sampling before feeding to the second layer for classification.
This procedure enables the network's learning to match the learning dynamics of analytic learning~\cite{zhuang2021blockwise, zhuang2022acil}.

\subsubsection{Re-training via AL}
Specifically, we feed the input samples $(X_{k-1,v}^{\text{train}}, X_{k-1, t}^{\text{train}})$ and the prompts ($P^v, P^t$) into the multi-modal backbone $\mathcal{F}$ to extract the joint embeddings (i.e., last layer hidden representations), and then up-sample followed by an activation function:
\begin{equation}
    H_{k-1} = f_{\sigma} (\mathcal{F}([P^t, X_{k-1, t}^{\text{train}}, P^v, X_{k-1,v}^{\text{train}}]) W_{\text{up}})
\end{equation}
where $W_{\text{up}}$ denotes the parameters of the first up-sampling layer, and $f_{\sigma}$ is the activation function.
Here, we follow other analytic works~\cite{zhuang2022acil, Zhuang2024} to adopt the random initialization for $W_{\text{up}}$ and \textit{ReLU} for $f_{\sigma}$.

We then map the up-sampled hidden representations into the label matrix $Y_{k-1}^{\text{train}}$ via the second linear classifier layer, whose weights can be computed by solving:
\begin{equation}
    \underset{W_{k-1}}{\text{argmin}} = \parallel Y_{k-1}^{\text{train}} - H_{k-1} W_{k-1} \parallel^{2}_{\textup{F}} + \eta \parallel W_{k-1} \parallel^{2}_{\textup{F}},
\end{equation}
where $\parallel\cdot\parallel_{\textup{F}}$ is the Frobenius form, and $\eta$ is the regularization term.
For the first task, the above equation can be denoted as:
\begin{equation}
\label{equation_phase0_1}
    \underset{W_{1}}{\text{argmin}} = \parallel Y_{1}^{\text{train}} - H_{1} W_{1} \parallel^{2}_{\textup{F}} + \eta \parallel W_{1} \parallel^{2}_{\textup{F}},
\end{equation}
The optimal estimation of $W_1$ can be computed in:
\begin{equation}
\label{equation_phase0_2}
    \hat{W}_{1} = (H_1^T H_{1} + \eta I)^{-1} H_1^T Y_{1}^{\text{train}},
\end{equation}

Next, assume that we are given $\mathcal{D}_{1}^{\text{train}},\dots,\mathcal{D}_{k-1}^{\text{train}}$, the learning problem using all seen data at step $k-1$ can be formulated as:
\begin{equation}
\label{equation_phasek}
\scalebox{1.0}{$
    \underset{W_{k-1}}{\text{argmin}} = \parallel Y_{1:k-1}^{\text{train}} -  H_{1:k-1} W_{k-1} \parallel^{2}_{\textup{F}} + \eta\parallel W_{k-1} \parallel^{2}_{\textup{F}},
    $}
\end{equation}
where \\
$
\scalebox{0.85}{
$
\begin{gathered}
Y_{1:k-1}^{\text{train}} = 
\begin{bmatrix}
    Y_{1}^{\text{train}} & 0 & 0 & \cdots & 0 \\
    0 & Y_{2}^{\text{train}} & 0 & \cdots & 0 \\
    & & \vdots & & \\
    0  & 0 & \cdots & 0 & Y_{k-1}^{\text{train}}
\end{bmatrix}
,
H_{1:k-1} =
\begin{bmatrix}
    H_{1} \\
    H_{2} \\
    \vdots \\
    H_{k-1}
\end{bmatrix}
\end{gathered}$}
$

The solution to (\ref{equation_phasek}) can be obtained as:
\begin{equation}
    \hat{W}_{k-1} = (H_{1:k-1}^T H_{1:k-1} + \eta I)^{-1} H_{1:k-1}^T Y_{1:k-1}^{\text{train}},
\end{equation}
where $\hat{W}_{k-1} \in \mathbb{R}^{d \times \sum_{i=1}^{k-1} d_{y_i}}$.

The goal of MMCIL is to sequentially learn new tasks on $\mathcal{D}_{k}^{\text{train}}$ given a network trained on $\mathcal{D}_{1:k-1}^{\text{train}}$.
% However, in above equation, we still need previous data.
% To mitigate this reliance, we re-formulate the CIL process as a RLS task indicated in the following theorem.
Nevertheless, the equation above shows that previous data is still required. 
To reduce this dependency, 
let $R_{k-1} = (H_{1:k-1}^T H_{1:k-1} + \eta I)^{-1}$, we redefine the MMCIL process as an RLS task as outlined in the subsequent theorem.

\noindent \textbf{Theorem 1.} At step $\mathcal{T}_k$, given its training data $D_{k}^{\text{train}}$ and the estimated weights of $\hat{W}_{k-1}$ of step $\mathcal{T}_{k-1}$, $\hat{W}_{k}$ can be recursively obtained by:
\begin{equation}
\label{equation:W}
\scalebox{1.0}{$
    \hat{W}_{k} = \hat{W}_{k-1} - R_k H_k^T H_{k} \hat{W}_{k-1} + R_k H_k^T Y_{k}^{\text{train}},
    $}
\end{equation}
where 
\begin{equation}
\label{equation:R}
\scalebox{1.0}{$
    R_{k} = R_{k-1} - R_{k-1} H_k^T(H_{k} R_{k-1} H_k^T + I)^{-1}H_{k}R_{k-1},
    $}
\end{equation}
\textit{proof}. See the Appendix.

Theorem 1 indicates that the weights of joint training can be recursively obtained by training on the data from $\mathcal{D}_{1}^{\text{train}}$ to $\mathcal{D}_{k}^{\text{train}}$ sequentially.
This implies that, by freezing the multi-modal backbone, the MMCIL becomes equivalent to its joint training counterpart, as demonstrated in the theorem.
In other words, the model trained incrementally yields the same weights as that trained on both current and all previous data.

\begin{algorithm}[!h]
\caption{The procedure of PAL}
% \SetAlgoLined
\begin{algorithmic}[1]
\REQUIRE Training data $\mathcal{D}^{\text{train}}_{1:K}$, Prompt Pool $\mathcal{P}_v$ and $\mathcal{P}_t$, multi-modal backbone $\mathcal{F}$, classifier $\mathcal{C}$, reconstruction weight $\lambda$, regularization weight $\eta$.
\STATE \textbf{Initial training via BP}: Train the prompt module and the classifier via BP on the data of the first task.
\STATE \textbf{Initial re-training via AL}: Obtain the initial weight $\hat{W}_{1}$ using Eq.~(\ref{equation_phase0_2}) and $R_1 = (H_1^T H_1 + \eta I)^{-1}$ using the data of the first task.
\FOR{k=2 to K (with $\mathcal{D}^{\text{train}}_{k}$, $\hat{W}_{k-1}$, and $R_{k-1}$)} 
    \STATE i) Update $\mathcal{P}_v$ and $\mathcal{P}_t$ via BP using Eq.~(\ref{eq:bp_train}). \\
    \STATE ii) Update $R_k$ and $\hat{W}_{k}$ via AL using Eq.~(\ref{equation:W}) and Eq.~(\ref{equation:R}), respectively.
\ENDFOR
\end{algorithmic}
\end{algorithm}

\section{Experiments}
\subsection{Datasets}
We conduct experiments on two widely used datasets, i.e., UPMC-Food101~\cite{Wang2015} and N24News~\cite{n24news}.
Specifically, UPMC-Food101 consists of noisy image-text paired data over 101 food categories and includes 61,142 training, 6,846 validation, and 22,716 test samples, respectively.
We follow the experimental protocol adopted in~\cite{zhao2024arxiv}, in which the category with the fewest samples are removed and 100 categories are remained.
For the N24News dataset, which is generated from New York Times with 24 categories and contains both text and image information in each news, contains 48,988 training, 6,123 validation, and 6,124 testing samples, respectively.

\begin{table*}[!t]
\begin{center}
\caption{Results of different approaches on the UPMC-Food101 dataset under various modality-missing scenarios. 
The evaluation metrics are \textit{Acc} and \textit{FG}. Missing modality occurs during both training and testing.
The bold part denotes the overall best results, and the underlined part denotes the best results of the compared baselines.}
\resizebox{\linewidth}{!}{
\begin{tabular}{ll|ccc|ccc|ccc|ccc|ccc}
\toprule \toprule
\multirow{3}{*}{Metric} & \multicolumn{1}{c|}{$\eta$} & \multicolumn{3}{c|}{10\%} & \multicolumn{3}{c|}{30\%} & \multicolumn{3}{c|}{50\%} & \multicolumn{3}{c|}{70\%} & \multicolumn{3}{c}{90\%} \\ \cmidrule{3-17}
& \multicolumn{1}{c|}{\#Image} & 100\%  & 90\%  & 95\%  & 100\% & 70\% & 85\% & 100\%  & 50\%  & 75\% & 100\%  & 30\%  & 65\%  & 100\% & 10\% & 55\%  \\
& \multicolumn{1}{c|}{\#Text} & 90\% & 100\%  & 95\%  & 70\% & 100\% & 85\% & 50\% & 100\% & 75\%  & 30\%  & 100\% & 65\%  & 10\% & 100\% & 55\%  \\ \midrule

\multirow{6}{*}{\textit{Acc}~($\uparrow$)} & MAP~\cite{Lee2023cvpr} & 20.66 & 21.53 & 22.84 & 16.57 & 24.00 & 20.66 & 18.18 & 23.85 & 18.66 & 17.68 & 22.48 & 20.00 & 16.92 & 24.89 & 18.41 \\ 
& MSP~\cite{Jang2024icassp} & 21.45 & 23.29 & 22.12 & 21.37 & 22.21 & 21.89 & 17.88 & 21.57 & 18.10 & 19.29 & 21.22 & 19.76 & 18.62 & 21.69 & 20.45 \\ 
& L2P~\cite{wang2022cvpr} & 34.09 & 35.21 & 34.00 & 30.40 & 34.75 & 29.76 & 28.95 & 32.18 & 25.30 & 26.57 & 30.43 & 24.62 & 24.94 & 29.77 & 24.62 \\ 
& DualPrompt~\cite{wang2022eccv} & 59.56 & 59.90 & 58.18 & 51.86 & 52.66 & 48.56 & 47.70 & 50.22 & 43.67 & 43.09 & 50.28 & 40.69 & 37.26 & 51.16 & 40.69 \\ 
& RebQ~\cite{zhao2024arxiv} & 68.67 & 72.46 & 71.06 & 62.06 & 71.62 & 66.37 & 55.87 & 69.23 & 62.40 & 50.00 & 69.41 & 59.92 & 48.15 & 67.71 & 54.67 \\ 
& ACIL~\cite{zhuang2022acil} & \underline{78.62} & \underline{79.78} & \underline{79.24} & \underline{73.31} & \underline{78.49} & \underline{74.84} & \underline{68.10} & \underline{77.49} & \underline{71.71} & \underline{63.84} & \underline{76.44} & \underline{68.35} & \underline{61.34} & \underline{76.93} & \underline{65.93} \\ 
& PAL & \textbf{82.45} & \textbf{84.66} & \textbf{83.50} & \textbf{76.80} & \textbf{83.09} & \textbf{79.14} & \textbf{71.92} & \textbf{81.56} & \textbf{75.76} & \textbf{66.78} & \textbf{80.51} & \textbf{72.59} & \textbf{64.34} & \textbf{80.68} & \textbf{69.69} \\ 
\midrule

\multirow{6}{*}{\textit{FG}~($\downarrow$)} & MAP~\cite{Lee2023cvpr} & 82.50 & 82.20 & 80.13 & 83.77 & 77.12 & 79.81 & 79.77 & 75.59 & 79.80 & 77.82 & 75.21 & 75.95 & 76.60 & 70.84 & 75.19 \\ 
& MSP~\cite{Jang2024icassp} & 80.03 & 78.98 & 79.35 & 79.95 & 76.02 & 79.01 & 77.31 & 77.99 & 78.04 & 78.17 & 75.21 & 76.54 & 75.14 & 71.23 & 74.31 \\ 
& L2P~\cite{wang2022cvpr} & 5.80 & 5.51 & 6.55 & 6.53 & 4.66 & 6.87 & 6.33 & \underline{4.55} & \underline{6.17} & \textbf{\underline{5.80}} & 3.99 & \underline{5.66} & \textbf{\underline{5.60}} & \underline{4.54} & 5.66 \\ 
& DualPrompt~\cite{wang2022eccv} & \textbf{\underline{3.38}} & \textbf{\underline{1.72}} & 6.32 & \underline{6.46} & \underline{5.11} & 8.73 & \underline{6.09} & 5.03 & 8.52 & 8.96 & 5.19 & 7.26 & 11.46 & 4.80 & 7.29 \\ 
& RebQ~\cite{zhao2024arxiv} & 9.50 & 7.64 & 8.22 & 10.76 & 6.44 & 8.07 & 12.07 & 5.84 & 8.24 & 12.47 & \underline{3.73} & 8.56 & 12.76 & 4.71 & 8.78 \\ 
& ACIL~\cite{zhuang2022acil} & 6.24 & 5.15 & \underline{5.83} & 7.76 & 5.57 & \underline{6.56} & 8.64 & 5.18 & 7.48 & 9.88 & 5.35 & 7.63 & 10.42 & 5.33 & \textbf{\underline{4.33}} \\
& PAL & 4.35 & 3.74 & \textbf{3.72} & \textbf{5.24} & \textbf{3.53} & \textbf{4.32} & \textbf{5.76} & \textbf{3.32} & \textbf{4.59} & 6.21 & \textbf{3.00} & \textbf{4.92} & 7.01 & \textbf{2.67} & 5.32 \\ 
\bottomrule \bottomrule
\end{tabular}}
\label{tab:food101_main}
\end{center}
\end{table*}

\begin{table*}[!t]
\begin{center}
\caption{Results of different approaches on the N24News dataset under various modality-missing scenarios. 
The evaluation metrics are \textit{Acc} and \textit{FG}. Missing modality occurs during both training and testing.
The bold part denotes the overall best results, and the underlined part denotes the best results of the compared baselines.}
\resizebox{\linewidth}{!}{
\begin{tabular}{ll|ccc|ccc|ccc|ccc|ccc}
\toprule \toprule
\multirow{3}{*}{Metric} & \multicolumn{1}{c|}{$\eta$} & \multicolumn{3}{c|}{10\%} & \multicolumn{3}{c|}{30\%} & \multicolumn{3}{c|}{50\%} & \multicolumn{3}{c|}{70\%} & \multicolumn{3}{c}{90\%} \\ \cmidrule{3-17}
& \multicolumn{1}{c|}{\#Image} & 100\%  & 90\%  & 95\%  & 100\% & 70\% & 85\% & 100\%  & 50\%  & 75\% & 100\%  & 30\%  & 65\%  & 100\% & 10\% & 55\%  \\
& \multicolumn{1}{c|}{\#Text} & 90\% & 100\%  & 95\%  & 70\% & 100\% & 85\% & 50\% & 100\% & 75\%  & 30\%  & 100\% & 65\%  & 10\% & 100\% & 55\%  \\  \midrule 

\multirow{6}{*}{\textit{Acc}~($\uparrow$)} & MAP~\cite{Lee2023cvpr} & 17.34 & 16.98 & 18.18 & 16.23 & 15.30 & 17.14 & 16.24 & 14.97 & 16.83 & 15.43 & 14.32 & 17.34 & 14.12 & 13.12 & 14.08 \\ 
& MSP~\cite{Jang2024icassp} & 19.12 & 20.69 & 21.13 & 19.34 & 18.21 & 19.37 & 17.91 & 18.76 & 18.01 & 17.13 & 16.15 & 17.24 & 16.88 & 15.19 & 14.78 \\ 
& L2P~\cite{wang2022cvpr} & 27.23 & 25.69 & 27.18 & 25.32 & 24.11 & 22.37 & 23.47 & 24.29 & 22.06 & 21.43 & 23.81 & 20.51 & 20.78 & 21.32 & 19.83 \\ 
& DualPrompt~\cite{wang2022eccv} & 36.67 & 34.07 & 35.98 & 34.29 & 33.76 & 34.19 & 32.24 & 33.01 & 32.36 & 30.48 & 31.91 & 30.73 & 29.94 & 30.13 & 29.27\\ 
& RebQ~\cite{zhao2024arxiv} & 42.37 & 42.14 & 44.19 & 41.66 & 39.75 & 39.44 & 40.48 & 38.42 & 38.38 & 37.44 & 36.84 & 37.27 & 38.06 & 35.44 & 34.56 \\ 
& ACIL~\cite{zhuang2022acil} & \underline{51.50} & \underline{50.97} & \underline{50.63} & \underline{48.86} & \underline{49.74} & \underline{48.46} & \underline{45.89} & \underline{47.36} & \underline{46.71} & \underline{43.28} & \underline{44.99} & \underline{43.74} & \underline{42.66} & \underline{43.30} & \underline{41.19} \\ 
& PAL & \textbf{54.66} & \textbf{54.68} & \textbf{54.52} & \textbf{51.31} & \textbf{52.82} & \textbf{50.37} & \textbf{48.76} & \textbf{50.50} & \textbf{48.05} & \textbf{45.34} & \textbf{48.15} & \textbf{46.02} & \textbf{45.50} & \textbf{47.57} & \textbf{44.03} \\ 
\midrule

\multirow{6}{*}{\textit{FG}~($\downarrow$)} & MAP~\cite{Lee2023cvpr} & 60.23 & 59.40 & 61.34 & 58.12 & 55.48 & 56.72 & 57.09 & 56.81 & 57.92 & 55.02 & 52.84 & 53.17 & 54.10 & 55.27 & 53.19 \\ 
& MSP~\cite{Jang2024icassp} & 59.11 & 55.39 & 57.18 & 56.89 & 54.45 & 55.72 & 57.62 & 53.37 & 55.76 & 53.14 & 52.57 & 53.27 & 51.03 & 52.89 & 51.38 \\ 
& L2P~\cite{wang2022cvpr} & 11.23 & 10.87 & 11.12 & 10.28 & 11.38 & 12.51 & 11.82 & 10.86 & 12.42 & 10.38 & 15.36 & 13.91 & \textbf{\underline{8.03}} & \underline{10.12} & 12.26 \\ 
& DualPrompt~\cite{wang2022eccv} & 10.42 & \underline{8.19} & \underline{9.93} & 10.41 & 10.02 & 10.93 & \underline{9.32} & 11.18 & 11.23 & \textbf{\underline{9.28}} & 13.12 & 12.61 & 13.04 & 11.82 & 13.51\\ 
& RebQ~\cite{zhao2024arxiv} & 12.92 & 13.26 & 14.02 & 12.17 & 13.37 & 11.85 & 15.48 & 15.40 & 14.86 & 16.63 & 17.44 & 14.64 & 14.08 & 15.05 & 15.93 \\ 
& ACIL~\cite{zhuang2022acil} & \underline{9.50} & 9.54 & 10.37 & \underline{10.20} & \underline{9.56} & \underline{10.53} & 10.57 & \underline{10.23} & \underline{10.17} & 11.21 & \underline{10.95} & \underline{10.29} & 10.39 & 10.64 & \underline{11.19}\\ 
& PAL & \textbf{8.18} & \textbf{6.62} & \textbf{8.07} & \textbf{8.17} & \textbf{7.56} & \textbf{8.89} & \textbf{7.02} & \textbf{7.01} & \textbf{6.98} & 9.92 & \textbf{8.75} & \textbf{8.66} & 8.27 & \textbf{8.20} & \textbf{7.23} \\ 
\bottomrule \bottomrule
\end{tabular}}
\label{tab:n24news_main}
\end{center}
\end{table*}

\subsection{Implementation Details}
\subsubsection{Training Details} Following ~\cite{zhao2024arxiv, Lee2023cvpr}, we adopt the pre-trained multimodal transformer ViLT~\cite{vilt2021} as our backbone, which stems from Vision Transformers~\cite{vit2021} and advances to process multimodal inputs with tokenized texts and patched images.
Without using modality-specific feature extractors, ViLT is pre-trained on several image-text datasets (e.g., MS-COCO~\cite{mscoco2014} and Visual Genome~\cite{vg2016}).

We freeze all the parameters of the pre-trained ViLT backbone and only train the learnable prompts and the linear classifier.
For the image, and text prompt pools, we set the length of prompts, the number of prompt layers, and the prompt pool size to 8, 8, 128, respectively.
The reconstruction weight is set to 0.01 and the regularization is set to 1.
We use the AdamW~\cite{adamw} to optimize the model with a learning rate of 1e-4 and batch size of 4.
All experiments are conducted using one Nvidia A100 GPU with the results averaged over 3 runs.

\subsubsection{Class-Incremental Protocol}
We follow the protocol adopted in previous works~\cite{zhao2024arxiv}, in which all the classes are evenly distributed over the incremental steps.
To be specific, we evenly divide the N24News dataset into 6 incremental steps, each of which contains 4 classes.
For the UPMC-Food101 dataset, we evenly split 100 categories into 10 sessions, each of which contains 10 categories.
Most existing methods only report small-phase results, e.g., those of $K=5, 10$, we include $K=25, 50$ as well to validate PAL's large-step performance.

\subsubsection{Setting of Missing Modality} 
We focus on the more challenging scenario where the missing modality can occur in both training and testing stages during the incremental steps, where each modality for each sample can be lost.
Following~\cite{zhao2024arxiv, Lee2023cvpr}, the missing-text (missing-image) with the missing rate $\eta\%$ indicates there are $\eta\%$ image-only (text-only) data and $(1-\eta)\%$ complete data, while for the missing-both case, there are $ \frac{\eta}{2}\%$ image-only data, $ \frac{\eta}{2}\%$ text-only data and $(1-\eta)\%$ complete data during each incremental step.

\subsection{Evaluation Metric}
We adopt two standard CIL metrics: Average Accuracy ($Acc$) and Average Forgetting ($FG$).
$Acc = \frac{1}{K} \sum_{k=1}^{K} a_{k,K}$, where $a_{i,j}$ indicates the test accuracy on the $i$-th task after training the model on the $j$-th task.
$FG = \frac{1}{K-1} \sum_{k=1}^{K-1} \text{max}_{z \in {t, \cdots, K-1} (a_{k,z} - a_{k, K})}$, reveals the degree to which a CIL method forgets the learned knowledge.

\begin{figure*}
\centering
\includegraphics[scale=1.5]{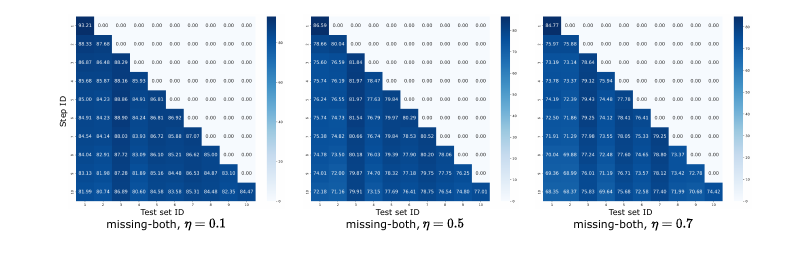}
\caption{Testing accuracy at each incremental step on the UPMC-Food101 dataset, when the missing rate $\eta$ is 0.1, 0.5 and 0.7.}
\label{fig:step_acc_food}
\end{figure*}

\begin{figure*}
\centering
\includegraphics[scale=1.5]{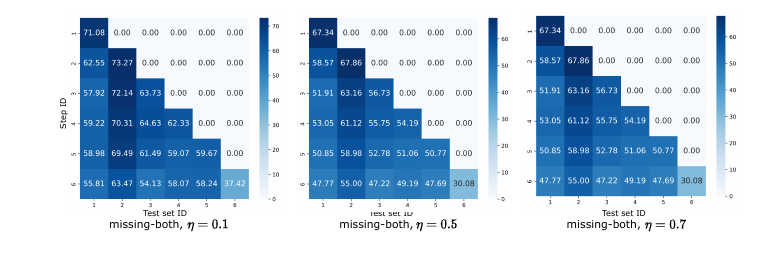}
\caption{Testing accuracy at each incremental step on the N24News dataset, when the missing rate $\eta$ is 0.1, 0.5 and 0.7.}
\label{fig:step_acc_news}
\end{figure*}

\section{Results and Analysis}
\subsection{Main results}
To demonstrate the effectiveness of the proposed PAL, we compare it against recent state-of-the-art methods, including MAP~\cite{Lee2023cvpr}, MSP~\cite{Jang2024icassp}, L2P~\cite{wang2022cvpr}, DualPrompt~\cite{wang2022eccv}, RebQ~\cite{zhao2024arxiv}, and ACIL~\cite{zhuang2022acil}.
Specifically, MAP and MSP are methods that utilize pre-trained multi-modal transformers to tackle the missing modality problem, but they are limited to static datasets.
L2P and DualPrompt are prompt-based CIL methods: L2P introduces a prompt pool mechanism for tuning pre-trained backbones, while DualPrompt leverages complementary task-invariant and task-specific prompts to further enhance the performance.
RebQ employs query-based optimization for prompt learning during MMCIL with missing modality.
In contrast, ACIL is an analytic learning-based CIL method that focuses on a single modality.

Table~\ref{tab:food101_main} and~\ref{tab:n24news_main} summarize the experimental results with different missing rates on UMPC-Food101 and N24News datasets, respectively.
We also present the accuracy at each incremental step on both datasets in Figure~\ref{fig:step_acc_food} and~\ref{fig:step_acc_news}.
As shown in the table, PAL outperforms all comparative baselines across various configurations in both Average Accuracy (\textit{Acc}) and Forgetting (\textit{FG}) by a large margin.
For instance, on the UMPC-Food101 dataset, PAL yields 64.34\%, 80.68\%, and 69.69\% for the missing-text, missing-image, and missing-both scenarios, respectively, when the total missing rate is 90\%.
While for the N24News dataset, PAL achieves 45.50\%, 47.57\%, and 44.03\%, compared to RebQ, which only achieves 38.06\%, 35.44\%, and 34.56\% for the missing-text, missing-image, and missing-both scenarios, respectively, when the total missing rate is 90\%.
Regarding the forgetting rate, PAL also achieves the lowest values across most settings.
For example, PAL obtains 5.76\% on missing-text, 3.32\% on missing-image, and 4.59\% on missing-both when the missing rate is 50\% on the UMPC-Food101 dataset.
Similarly, on the N24News dataset, PAL shows 7.02\% on missing-text, 7.01\% on missing-image, and 6.98\% on missing-both when the missing rate is 50\%.
Compared with RebQ, PAL leverages analytic learning to mitigate the forgetting issue, leading to \textit{FG} values as low as 5.32\% in the 90\% missing-both scenario, significantly outperforming 8.78\% of RebQ on UMPC-Food101.
Compared with ACIL, PAL equips with prompt learning to adapt to new information, leading to 69.69\% accuracy in the 90\% missing-both scenario, outperforming 65.93\% of ACIL by a large margin, with a slight forgetting increase from 4.33\% to 5.32\%.
This indicates that PAL, by unifying prompt tuning and analytic learning, is able to tackle general miss-modality cases and provide good instruction for model prediction during the multi-modal CIL steps with missing modality.

% \begin{figure}
%     \centering
%     \includegraphics[scale=0.3]{figures/acc.png}
%     \caption{Testing accuracy at each incremental step on the UPMC-Food101 dataset. The scenarios are missing-both cases with 70\% missing rate.}
%     \label{fig:each_step}
% \end{figure}

% In Figure~\ref{fig:each_step}, we present the testing accuracy at each incremental step of our PAL and other baselines on the UPMC-Food101 dataset under the missing-both cases with 70\% missing rate.
% It can be observed that PAL achieves the best performance at each incremental step, while other compared methods suffer from catastrophic forgetting, leading to severe performance drops.

\subsection{Effectiveness of Prompt Module}
In this subsection, we first analyze the effect of the proposed prompt module, i.e., modality-specific prompt pools.
We conduct ablation experiments under the missing-both cases with 70\% missing rate on the UMPC-Food101 dataset as the default scenarios if not specified otherwise.

To show the effectiveness of the proposed prompt module $\mathcal{P}$, firstly, we remove it and only train with the AL module.
Then, we replace the modality-specific pools (i.e., $\mathcal{P}_v$ and $\mathcal{P}_t$) with a single share prompt pool.
Finally, we introduce a variant ``w/ prompt vector", relying on a singular prompt vector instead of a pool.
As shown in Table~\ref{tab:ablation_pool}, the performance has a significant drop (i.e., 68.36/7.82 vs 72.59/4.92) if we remove the prompt module, suggesting that the prompt module could preserve the holistic representation of the missing-modality data for the following analytic CIL procedure.
% If we remove the AL module, the performance will decrease drastically from 72.59/4.92 to 59.94/8.97.
Similarly, the single prompt pool and the vector suffer from capturing knowledge with each modality and entanglement between different modalities, the modality-specific prompt pool could provide better discriminative knowledge between tasks.

Moreover, the performance is also influenced by various prompting strategies, including the layer at which prompts are inserted, the length of the prompts, and the size of the prompt pool. 
Next, we will study the effects of these factors.

\begin{table}[]
\centering
\caption{Ablation studies of the prompt module on the UPMC-Food101 dataset.}
\begin{tabular}{lcc}
\toprule \toprule
Method & \textit{Acc}~($\uparrow$) & \textit{FG}~($\downarrow$) \\ \midrule
PAL                         & 72.59  & 4.92 \\ \midrule
w/o prompt module           & 68.36  & 7.82 \\
w/o AL module               & 59.94  & 8.97 \\ \midrule
w/ single prompt pool       & 70.70   & 8.19     \\
w/ prompt vector            & 67.92  & 7.60 \\
\bottomrule \bottomrule
\end{tabular}
\label{tab:ablation_pool}
\end{table}

\subsubsection{Position of Prompts} 
\begin{figure}
    \centering
    \includegraphics[scale=1.3]{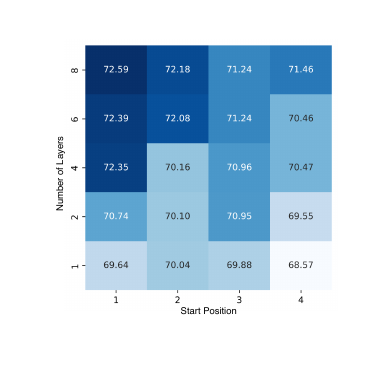}
    \caption{Ablation study on the location of prompting layers. }
    \label{fig:position}
\end{figure}

We conduct ablation experiments to analyze the effect of different layers to attach the prompting vectors in Figure~\ref{fig:position}.
We observe that the performance increases intuitively as the number of layers increases, while a more critical factor is which layer to start attaching prompts, i.e., the earlier layer the better.
This finding suggests that early layers with prompts have a more significant influence on the model performance, which is possibly due to that multimodal inputs are fused from the beginning of the transformer, in which the degree of fusion increases when the layer is deeper.
Therefore, it is more effective for early layers to be guided by the instruction of modality-specific prompts.

\begin{figure}
    \centering
    \includegraphics[scale=1.6]{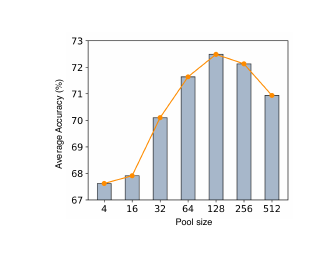}
    \caption{Model performance comparison when we vary the prompt pool size.}
    \label{fig:prompt_size}
\end{figure}

\begin{figure}
    \centering
    \includegraphics[scale=1.6]{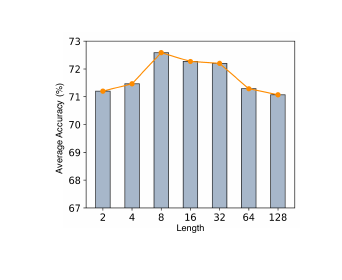}
    \caption{Model performance comparison when we vary the prompt length.}
    \label{fig:prompt_length}
\end{figure}

\begin{figure}
    \centering
    \includegraphics[scale=1.0]{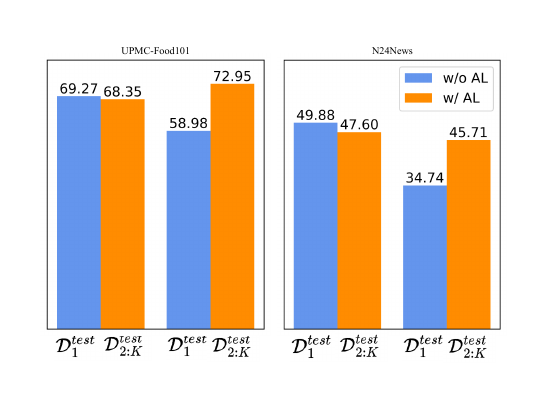}
    \caption{The trade-off between stability and plasticity via AL module.}
    \label{fig:trade_off}
\end{figure}

\subsubsection{Pool Size and Prompt Length}
We study the effect of the prompt pool size and prompt length in Figure~\ref{fig:prompt_size} and~\ref{fig:prompt_length}, respectively.
Figure~\ref{fig:prompt_size} compares the model performance across different pool sizes, revealing that smaller pools are still prone to catastrophic forgetting.
The performance improves up to the pool size of 128, beyond which further enlargement does not yield improvement.

Figure~\ref{fig:prompt_length} illustrates the impact of varying prompt length on model performance in terms of accuracy.
The performance improves as the prompt length increases from 2 to 8, reaching the highest accuracy of 72.59\% at a length of 8.
Beyond this, the accuracy begins to plateau and then gradually decreases for long prompts.
This suggests that overly short prompts may fail to capture sufficient task-relevant information, while excessively long prompts might introduce noise or redundancy, hindering model performance.

\subsection{Effectiveness of AL Module}
In this subsection, we study the impact of the proposed AL module.
To benchmark its importance, we conduct experiments by training the model without the AL module.
As indicated in Table~\ref{tab:ablation_pool}, the AL module plays a vital role in against the issue of catastrophic forgetting during multi-modal CIL phases (i.e., 59.94/8.97 vs 72.59/4.92).
The performance of the AL module is primarily influenced by the expansion size and regularization weight.

\subsubsection{Up-sampling size and Regularization Weight}
As shown in Table~\ref{tab:expansion_reg}, a large feature up-sampling size helps to capture more discriminative information, while too large (i.e., 20k) will degrade the performance.
For the regularization weight, it plays an important role but behaves robustly during MMCIL experiments.
As shown in rows 2-4 in Table~\ref{tab:expansion_reg}, PAL yields similar performance for values from 0.5 to 1.0, however, PAL could suffer from a strong accuracy reduction for too small values (e.g., 0.1).

\subsubsection{Trade-off between Stability and Plasticity}
Balancing stability ( retaining old knowledge) and plasticity (acquiring new knowledge) is a core challenge in CIL.
Through the AL module, our PAL can effectively manage a more reasonable stability-plasticity balance.
To illustrate this, we report the \textit{Acc} separately for the first task (reflecting stability) and the remaining tasks (indicating plasticity).
As depicted in Figure~\ref{fig:trade_off}, the AL module enables substantial improvements in novel class accuracy (plasticity), significantly outweighing the minimal reduction in old class accuracy (stability). For instance, the AL module boosts the novel class accuracy by an impressive 13.97\% (from 58.98\% to 72.95\%), while the old class accuracy experiences only a minor decrease of 0.92\% (from 69.29\% to 68.35\%). 
This demonstrates PAL's capability to enhance incremental learning performance by achieving a more balanced trade-off between stability and plasticity.

\begin{table}[]
\caption{Ablation study of up-sampling size and regularization weight.}
\resizebox{\linewidth}{!}{
\begin{tabular}{c|ccc|cc}
\toprule \toprule
\multirow{2}{*}{Up-sampling size} & \multicolumn{3}{c|}{w/ regularization} & \multirow{2}{*}{\textit{Acc}$(\%) \uparrow$} & \multirow{2}{*}{\textit{FG}(\%) $\downarrow$} \\ \cmidrule{2-4}  & 0.1   & 0.5  & 1.0   &    &    \\ \midrule
8k   &  $\times$  & $\times$  & $\checkmark$  & 71.85 & 6.50     \\ 
15k  &  $\times$ & $\times$  & $\checkmark$   & 72.59 & 4.92 \\
15k  &  $\times$  & $\checkmark$  & $\times$  & 72.45 & 4.00 \\
15k  &  $\checkmark$  & $\times$  & $\times$  & 67.54   & 5.29 \\
20k  &  $\times$  & $\times$  & $\checkmark$   & 72.39   & 4.43 \\  
\bottomrule \bottomrule
\end{tabular}}
\label{tab:expansion_reg}
\end{table}

\begin{table}[]
\centering
\caption{The evolution of Average Accuracy \textit{Acc} and Forgetting \textit{FG} with the growing incremental step on the UPMC-Food101 dataset.}
\begin{tabular}{c|cc}
\toprule \toprule
Num. Step & \textit{Acc} $(\%) \uparrow$   &  \textit{FG} (\%) $\downarrow$ \\ \hline
5          & 73.19  & 4.13    \\ 
10         & 72.59  & 4.92    \\ 
20         & 72.46  & 5.84    \\ 
50         & 72.16  & 5.36    \\ 
\bottomrule \bottomrule
\end{tabular}
\label{tab:large-phase-acc}
\end{table}

\subsection{Large-step Performance}
We have demonstrated that the PAL exhibits a step-invariant property for MMCIL.
To empirically validate this claim, we conduct large-step incremental experiments on the UPMC-Food101 dataset, varying the number of tasks $K$ from 5 to 50. 
The network is trained incrementally on 100 classes, distributed evenly across $K$ steps. For instance, in the 10-task setup, 10 classes are added per step, while in the extreme case of $K=50$, only 2 classes are added at each step. 
As presented in Table~\ref{tab:large-phase-acc}, the average accuracy (\textit{Acc}) and forgetting (\textit{FG}) remain remarkably consistent across all scenarios. 
This demonstrates PAL's robustness and scalability, effectively handling fine-grained incremental steps with a slight decline in performance.

\section{Conclusion}
In this paper, we achieve the first attempt to tackle the challenging MMCIL with missing modality.
Specifically, PAL comprises a prompt module that leverages modality-specific prompts to preserve the comprehensive representation of the miss-modality data, along with an AL module that redefines the MMCIL task through an RLS solution.
These two modules can complement each other to alleviate the under-fitting issue stemming from the static backbone within the AL module, yielding discriminative representations and improved incremental performance, despite various missing-modality scenarios.
Experimental results show that PAL significantly outperforms state-of-the-art methods with 72.59\% and 46.02\% accuracy on UPMC-Food101 and N24News datasets respectively under the missing rate of 70\% in contrast to competitive baselines.
In the future, we plan to extend PAL for tri-modal learning and beyond.

\bibliographystyle{IEEEtran}
\bibliography{references}

\clearpage
\appendix
\section*{Proof of Theorem 1}
Here we give the proof of Theorem 1 in the main paper: \\
\setcounter{equation}{0}

\noindent Specifically, at step $k-1$, we have
\begin{align}
\label{eq_w_k-1}
    \hat{W}_{k-1} = (H_{1:k-1}^{T} H_{1:k-1} + \eta I)^{-1} (H_{1:k-1})^{T} Y^{\text{train}}_{1:k-1},
\end{align}
Hence, at step $k$ we then have 
\begin{align}
\label{eq_w_k}
    \hat{W}_{k} = (H_{1:k}^{T} H_{1:k} + \eta I)^{-1} (H_{1:k})^{T} Y^{\text{train}}_{1:k},
\end{align}
In the paper, we have defined
\begin{align}
\label{eq_r_m_k-1}
    R_{k-1} = (H_{1:k-1}^T H_{1:k-1} + \eta I)^{-1},
\end{align}
Here we also define, 
\begin{align}
\label{eq_Q_k_1}
    Q_{k-1} = H_{1:k-1}^{T} Y^{\text{train}}_{1:k-1},
\end{align}
Thus we can rewrite \eqref{eq_w_k-1} as
\begin{align}\label{xx}
    \hat{W}_{k-1} = R_{k-1} Q_{k-1},
\end{align}
Therefore, at step ${k}$ we have
\begin{align}\label{compact}
    \hat{W}_{k} = R_{k} Q_{k},
\end{align}
From \eqref{eq_r_m_k-1}, we can recursively calculate $R_{k}$ from $R_{k-1}$, i.e.,
\begin{align}
\label{eq_r_m_k}
    R_{k} = (R_{k-1}^{-1} + H_{k}^T H_{k})^{-1},   
\end{align}
According to the Woodbury matrix identity, we have
\begin{align}
    (A + UCV)^{-1} = A^{-1} - A^{-1} U (VA^{-1}U + C^{-1})VA^{-1},
\end{align}
Let $A = (R_{k-1})^{-1}$, $U = H_{k}^T$, $C = I$, $V = H_{k}$, in \eqref{eq_r_m_k}, we have
\begin{align}
\label{eq_R_update1}
    R_{k} = R_{k-1} - R_{k-1} H_{k}^T (H_{k} R_{k-1} H_{k}^T + I)^{-1}H_{k} R_{k-1},
\end{align}
Hence, $R_{k}$ can be recursively updated using its last-step counterpart $R_{k-1}$ and data from the current step (i.e., $H_{k}$). This proves the recursive calculation of $R_{k}$. \\
		
Next, we derive the recursive calculation of $\hat{W}_{k}$. 
To this end, we first recursively calculate $Q_{k}$, i.e.,
\begin{align}
\label{eq_R_update3}
    Q_{k} = H_{1:k}^{T} Y^{\text{train}}_{1:k} = Q_{k-1} + H_{k}^T Y^{\text{train}}_{k}
\end{align}
Let $K_k = (H_{k} R_{k-1} H_{k}^T + I)^{-1}$. Since 
\begin{align}
    I = K_k K^{-1}_k = K_k (H_{k} R_{k-1} H_{k}^T + I)
\end{align}
we have
$K_k = I - K_k H_{k} R_{k-1} H_{k}^T$.
Therefore,         
\begin{align}\label{a}
    & R_{k-1} H_{k}^T (H_{k} R_{k-1} H_{k}^T + I)^{-1}
    \notag \\
    &= R_{k-1} H_{k}^T K_k
    \notag \\
    &= R_{k-1} H_{k}^T (I - K_k H_{k} R_{k-1} H_{k}^T)
    \notag \\
    &= (R_{k-1} - R_{k-1} H_{k}^T K_k H_{k} R_{k-1}) H_{k}^T
    \notag \\
    &= R_{k} H_{k}^T,
\end{align}
Hence, $\hat{W}_{k}$ can be rewritten as
\begin{align}\nonumber
    \hat{W}_{k}
    &= R_{k} Q_{k} \\ \nonumber
    &= R_{k} (Q_{k-1} + H_{k}^T Y^{\text{train}}_{k}) \\ \label{eq_W_k_33}
    &= R_{k} Q_{k-1} + R_{k} H_{k}^T Y^{\text{train}}_{k},
\end{align}
By substituting \eqref{eq_R_update1} into $R_{k} Q_{k-1}$, we have
\begin{align}\nonumber
    & R_{k} Q_{k-1} = R_{k-1} Q_{k-1} \\ \nonumber
    & - R_{k-1} H_{k}^T (H_{k} R_{k-1} H_{k}^T + I)^{-1} H_{k} R_{k-1} Q_{k-1}\\ \label{eq_W_k_2}
    & = \hat{W}_{k-1} - R_{k-1} H_{k}^T (H_{k} R_{k-1} H_{k}^T + I)^{-1} H_{k} \hat{W}_{k-1},
\end{align}
According to \eqref{a}, \eqref{eq_W_k_2} can be rewritten as 
\begin{align}
\label{eq_RQprime}
    R_{k} Q_{k-1} = \hat{W}_{k-1} - R_{k} H_{k}^T H_{k} \hat{W}_{k-1},
\end{align}
By inserting \eqref{eq_RQprime} into \eqref{eq_W_k_33}, we have
\begin{align*}
    \hat{W}_{k} = \hat{W}_{k-1} - R_{k} H_{k}^T H_{k} \hat{W}_{k-1} + R_{k} H_{k}^T Y^{\text{train}}_{k},
\end{align*}
which proves the recursive calculation of $\hat{W}_{k}$.
Overall, now we have completed the proof of Theorem 1.

\end{document}